\documentclass[letterpaper]{article}
\usepackage{iccc}

\usepackage{times}
\usepackage{helvet}
\usepackage{courier}
\usepackage{graphicx}

\usepackage[dvipsnames]{xcolor}

\newcommand{\nameingr}[0]{\textsc{Name$\rightarrow$Ingr}}
\newcommand{\nameingrinst}[0]{\textsc{Name+Ingr$\rightarrow$Inst}}

\newcommand{\question}[1]{}
\newcommand{\note}[1]{}
\newcommand{\outline}[1]{}

\renewcommand{\question}[1]{\textcolor{red}{#1}}
\renewcommand{\note}[1]{\textcolor{blue}{#1}}
\renewcommand{\outline}[1]{\textcolor{green}{#1}}

\newcommand{\remove}[1]{\textcolor{red}{#1}}
\newcommand{\new}[1]{\textcolor{blue}{#1}}
\renewcommand{\remove}[1]{}
\renewcommand{\new}[1]{#1}

\title{Monte Carlo Tree Search for Recipe Generation using \mbox{GPT-2}}

\author{
Karan Taneja$^{1,2}$, Richard Segal$^2$, Richard Goodwin$^2$ \\
$^1$ School of Interactive Computing, Georgia Institute of Technology, GA, US\\
$^2$ Computational Creativity Group, IBM Research, NY, US\\
karantaneja@gatech.edu, rsegal@us.ibm.com, rgoodwin@us.ibm.com\\
}

\begin{document} 
\maketitle
\begin{abstract}
\begin{quote}

Automatic food recipe generation methods provide a creative tool for chefs to explore and to create new, and interesting culinary delights.
Given the recent success of large language models (LLMs), they have the potential to create new recipes that can meet individual preferences, dietary constraints, and adapt to what is in your refrigerator.
Existing research on using LLMs to generate recipes has shown that LLMs can be fine-tuned to generate realistic-sounding recipes.
However, on close examination, these generated recipes often fail to meet basic requirements like including chicken as an ingredient in chicken dishes.
In this paper, we propose \mbox{RecipeMC}, a text generation method using \mbox{GPT-2} that relies on Monte Carlo Tree Search (MCTS).
\mbox{RecipeMC} allows us to define reward functions to put soft constraints on text generation and thus improve the credibility of the generated recipes.
Our results show that human evaluators prefer recipes generated with \mbox{RecipeMC} more often than recipes generated with other baseline methods when compared with real recipes.

\end{quote}
\end{abstract}

\section{Introduction}

With the vast number of cooking recipes available online and the success of large language models (LLMs) such as \mbox{GPT-2} \cite{Radford2019LanguageLearners}, researchers have investigated the automatic generation of recipes by fine-tuning LLMs on large datasets of food recipes \cite{Bie2020RecipeNLG:Generation,Lee2020RecipeGPT:System}.
Automatic food recipe generation can be used in the creative process of recipe design where a chef can explore ingredient combinations, take inspiration for new recipes, write recipe drafts, or learn about flavor patterns.
LLMs trained to generate recipes can help chefs by generating multiple possible recipes, completing incomplete ingredient lists and recipe instructions.

Large language models such as \mbox{GPT-2} can be fine-tuned with large recipe datasets such as Recipe1M+ \cite{Marin2021Recipe1M+:Images} and \mbox{RecipeNLG} \cite{Bie2020RecipeNLG:Generation} to generate reasonable-looking recipes.
However, the quality of the output recipes is often limited due to the presence of repetitive text and inconsistencies between different parts of the recipe.
There are two main reasons for this:
(i) LLMs generate text by producing one token at a time and appending it to the existing text, resulting in a high focus on local coherence but a lack of attention to the long-term vision of the given context,
and (ii) fine-tuned models are typically trained on smaller datasets compared to the original model, which can lead to subpar generalization over the target domain.

\begin{figure*}[ht]
  \centering
  \renewcommand{\arraystretch}{1.2}
  \begin{tabular}{p{0.22\textwidth}p{0.70\textwidth}}
  \hline
  \textbf{Recipe Name} & John And Sarah’s Best Sausage, \textcolor{red}{\textbf{Shrimp}}, Ham And Chicken Jambalaya \\
  \hline
  \textbf{Ingredients (baseline)} & 4 celery ribs, chopped; 3-4 lbs chicken thighs; 1 teaspoon black pepper; 3 cups long grain white rice; 2 tablespoons garlic, chopped; 6 bay leaves; 1 teaspoon cayenne; 2 teaspoons salt; 1 lb smoked sausage; 1 teaspoon dried thyme; 1 cup green bell pepper, chopped; 1 cup ham, diced; 1 cup sweet onion, chopped; 2 cups chicken stock or 2 cups chicken broth; 4 cups water; 1 1/2 cups scallions, chopped; 8 tablespoons vegetable oil; 1 teaspoon oregano; 3 cups tomatoes, chopped (2 lb); 1 teaspoon white pepper \\
  \textbf{Ingredients (\mbox{RecipeMC})} & 1 teaspoon dried oregano; 2 cups long grain white rice; 2 teaspoons cajun seasoning; 1 teaspoon dried basil; 1 teaspoon cayenne; 1/2 lb smoked ham, cut into 1/4 inch dice; 1 lb chicken, boneless, skinless, cut into 1/2 inch dice; 1 teaspoon dried thyme; 1 lb andouille sausage, cut into 1/4 inch slices; 1/2 lb \textcolor{red}{\textbf{shrimp}}, shelled, deveined and cut into 1/4 inch dice; 2 tablespoons olive oil; 2 teaspoons salt; 1 large onion, chopped; 4 cups chicken broth; 2 garlic cloves, minced \\
  \hline
  \end{tabular}
  
  \vspace{10pt}
  \renewcommand{\arraystretch}{1.1}
  \begin{tabular}{p{0.22\textwidth}p{0.70\textwidth}}
  \hline
  \textbf{Recipe Name} & Crab-Stuffed Corn Muffins Casserole \\
  \textbf{Ingredients} & 1 1/2 tablespoons mayonnaise; 1 tablespoon minced \underline{green onion}; cooking spray; 1 cup cornmeal; 1/4 cup \underline{ grated pepper Jack cheese}; 1/2 teaspoon baking soda; 1/2 cup \underline{grated pepper Jack cheese}; 1/4 teaspoon Worcestershire sauce; 1/2 cup melted butter; 1 cup buttermilk; 8 ounces cooked crabmeat; 2 eggs; salt to taste; 1 cup all-purpose flour; 1 teaspoon Asian chili paste (sambal); 1 teaspoon fresh grated \textcolor{red}{\textbf{lemon zest}}; 1/2 teaspoon salt \\
  \hline
  \textbf{Instructions (baseline)} & Preheat oven to 350 degrees F (175 degrees C). Spray a 12-cup muffin tin with cooking spray. Whisk together the eggs, buttermilk, and melted butter in a bowl. Stir in 1 cup of the flour, 1/2 of the cornmeal, the baking soda, 1/4 of the salt, the \underline{grated cheese}, \underline{onion}, mayonnaise, chili paste, and Worcestershire sauce. Mix in the crabmeat [...] \\
  \textbf{Instructions (\mbox{RecipeMC})} & Preheat the oven to 400 degrees F (200 degrees C). Spray 12 muffin cups with cooking spray. Combine cornmeal, flour, baking soda, and salt in a large bowl. Whisk together buttermilk, butter, eggs, \underline{pepper Jack cheese}, mayonnaise, chili paste, \underline{green onion}, \textcolor{red}{\textbf{lemon zest}}, and Worcestershire sauce in a separate bowl. Stir into the dry ingredients [...] \\
  \hline
  \end{tabular}
  \caption{Examples of inconsistencies in generated recipes.
  In the first example, an important ingredient ``Shrimp'' is missing from the baseline ingredients list.
  In the second example, baseline instructions do not use the ingredient ``lemon zest''.
  Our method, RecipeMC, also refers to the complete names of the ingredients such as ``pepper Jack cheese'' and ``green onion'' rather than shortening them to ``grated cheese'' and  ``onion'' respectively.
  }
  \label{fig:InconsitentRecipes}
\end{figure*}

In this paper, we propose a method to sample from fine-tuned LLMs using Monte Carlo Tree Search (MCTS) and simple reward functions that put soft constraints on text generation.
These constraints aim to eliminate the irregularities in generated recipes, improving their plausibility and making them more appealing to human evaluators.
Our method does not require any additional training after the domain-specific fine-tuning of LLMs and can easily be wrapped over an API that exposes the next token probabilities.
Our work takes inspiration from \citeauthor{Chaffin2022PPL-MCTS:Decoding}~\shortcite{Chaffin2022PPL-MCTS:Decoding} where a discriminator network is used as the reward function with MCTS to generate text conditioned on specific classes.

Figure \ref{fig:InconsitentRecipes} shows a real example of (i) an ingredients list generated from a given recipe name and (ii) recipe instructions generated from a recipe name and ingredients list.
In the first example, while the recipe name mentions ``Shrimp,'' the baseline method fails to add shrimp to the ingredients list. In contrast, our proposed method, \mbox{RecipeMC}, includes shrimp in the ingredients list as expected.
In the second example, the baseline method does not employ ``lemon zest'' in recipe instructions while \mbox{RecipeMC} uses all the given ingredients as anticipated.
Further, \mbox{RecipeMC} also meticulously generates the complete names ``pepper Jack cheese'' and ``green onion'' unlike the baseline method which used the shortened versions ``grated cheese'' and ``onion''.

This paper makes three main contributions:
\begin{enumerate}
    \item 
    We introduce RecipeMC, a text-generation process based on MCTS that controls the output text with simple manually-defined reward functions to softly constrain recipes generated by a fine-tuned LLM model. 
    \item 
    We provide evidence that MCTS outperforms traditional sampling methods for recipe generation using common automatic evaluation metrics.
    \item 
    We conduct human evaluations that show that our generated recipes are often indistinguishable from human-created recipes and humans may prefer recipes generated by RecipeMC.
 \end{enumerate}

Next, we review the related literature to place our work in the context of previous research in food analysis and recipe generation.
We then describe our proposed recipe generation method  \mbox {RecipeMC} including \mbox{GPT-2} fine-tuning, MCTS, and the reward functions.
This is followed by a section on experiments and results including the automatic evaluation of \mbox{RecipeMC} with three other baseline methods and a human evaluation study.
Finally, we wrap up with a discussion, our ideas for future work, and concluding remarks.

\section{Related Literature}

\textbf{Food and Recipe Analysis:}
Salvador et al. \shortcite{Salvador2017LearningImages} introduced the Recipe1M dataset containing over 1M recipes with 800K food images.
With these recipes and corresponding food images, they trained text and image models to generate embeddings in a joint embedding space.
These text and image models with their common embedding space were used to retrieve recipes from food images (im2recipe retrieval task) by matching the embedding of a given image with those of recipes in an existing database.
Marin et al. \shortcite{Marin2021Recipe1M+:Images} extended this dataset to \mbox{Recipe1M+} by further adding 13M food images.
Min et al. \shortcite{Min2017BeingExploration} proposed a Multi-Modal Multi-Task Deep Belief Network (M3TDBN) to learn from multi-modal content and multi-attribute information in the food domain for cuisine classification, recipe image retrieval, and ingredient and attribute inference from food images.
Herranz, Min, and Jiang \shortcite{Herranz2018FoodKnowledge} reviewed work involving multiple modalities in food analysis including text, images, location, and cuisine.
Our paper focuses on coherent recipe text generation beginning with a recipe name to create ingredient lists and instructions.

\textbf{Recipe Generation:}
Chef Watson \cite{Varshney2019ACreativity} was based on a Bayesian model over a knowledge-representation schema containing culinary information such as relations between ingredients, geolocations, and chemical composition of ingredients in terms of flavor compounds.
Chef Watson could generate creative recipes whose quality was verified by professional chefs.
Wang et al. \shortcite{Wang2020Structure-AwareImages} proposed a method to generate recipes from food images using the Recipe1M dataset based on unsupervised extraction of paragraph structures and generating tree structures from images for a structure-aware generation.
Bi\'e et al. \shortcite{Bie2020RecipeNLG:Generation} introduced the RecipeNLG dataset for recipe generation which was created by expanding the Recipe1M recipes with over one million new recipes.
Lee et al. \shortcite{Lee2020RecipeGPT:System} introduced RecipeGPT, a GPT-2 based recipe generation model and evaluation system with a user interface for examining the quality and encouraging experimentation.
Reusch et al. \shortcite{Reusch2021RecipeGM:Model} introduced RecipeGM which generated recipes from a given list of ingredients (without quantities) using a hierarchical self-attention-based sequence-to-sequence model.
This model was proposed by \citeauthor{Fan2018HierarchicalGeneration}~\shortcite{Fan2018HierarchicalGeneration} for story generation where long-range dependencies are an important challenge.
Overall, RecipeGPT performs consistently better than RecipeGM except at n-gram repetition, but note that RecipeGPT is a much bigger model as compared to RecipeGM.
Antognini et al. \shortcite{Antognini2022AssistiveCritiquing} developed a method to edit recipes through critiques where the latent representation of the recipe is modified using its gradient with respect to the critique.
The critique is a list of desired ingredients in a recipe beginning with an initial recipe.
In this paper, we propose RecipeMC which uses GPT-2 model for recipe generation with soft constraints on ingredient lists and instructions for improving the coherence and overall quality of the recipes.
Our baseline methods use the same model as RecipeGPT, but use top-$p$ sampling instead of top-$k$ since it has been shown to consistently generate higher quality and more diverse text \cite{Holtzman2020TheDeGeneration}.
We show that RecipeMC consistently performs better than these baselines on all metrics and human evaluation.

\textbf{Constrained Text Generation:}
\mbox{LLMs}, such as \mbox{GPT-2}, struggle with maintaining the context of the prompt when generating structured responses.  Researchers have explored constrained text generation using different methods.
Zhang et al. \shortcite{Zhang2020POINTER:Pre-training} and Hsieh, Lee, and Lim \shortcite{Hsieh2021ENCONTER:Transformer} propose insertion-based transformer models to impose hard constraints that ensure the inclusion of given entities in the output text.
To impose these constraints, these models progressively add tokens between the given entity tokens to generate text but also inadvertently constrain the order in which tokens are produced.
Chaffin, Claveau, and Kijak \shortcite{Chaffin2022PPL-MCTS:Decoding} propose a method for controlling generation using MCTS and a discriminator model as the reward function to generate text conditioned on a given discriminator class.
This imposes a soft constraint on the output text, unlike insertion-based methods that enforce hard constraints.
Our work takes inspiration from this paper but uses simpler manually-defined reward functions instead of a discriminator model.
We linearly combine these reward functions to impose several soft constraints that encourage coherence and text quality in the recipes.
This can be used in future research to allow human-in-the-loop collaborative recipe generation where humans prescribe the reward functions.

\section{Recipe Generation}

\mbox{GPT-2} \cite{Radford2019LanguageLearners} is an LLM based on the transformer architecture \cite{Vaswani2017AttentionNeed}
and can generate text conditioned on an initial prompt using next token prediction.
\mbox{GPT-2} was trained on a large corpus with millions of web pages.
Fine-tuning a pre-trained LLM for specific tasks has been shown to have a significant advantage over training a model from scratch \cite{Radford2018ImprovingPre-Training}.
For our method, \mbox{RecipeMC}, we fine-tuned a \mbox{GPT-2} model using a large corpus of recipes collected from the internet.
Each recipe contains three components: recipe name, ingredients list, and recipe instructions.  Figure~\ref{fig:recipe-format} shows the recipe format we use to fine-tune our language model.
Here, \texttt{<|startofname|>} and \texttt{<|endofname|>} are special tokens used to denote the start and end of the recipe name respectively.
Similar tags are also used for denoting the ingredient and instruction sections.
We fine-tuned the \mbox{GPT-2} model using the \mbox{RecipeNLG} dataset \cite{Bie2020RecipeNLG:Generation} which contains over 2.1 million recipes.
We cleaned the dataset to remove text containing unwanted information such as text advertisements, empty or short ingredients, and instructions.

Similar to \citeauthor{Lee2020RecipeGPT:System}~\shortcite{Lee2020RecipeGPT:System}, we also experimented with a multi-task learning setup where a model is trained to generate multiple possible orderings of name, ingredients, and instructions instead of only the default name-ingredient-instruction ordering.
We did not see a decrease in perplexity over the test set, but rather a small increase, and decided to use the simpler single-order setup described in Figure \ref{fig:recipe-format}.

\begin{figure}[t]
\begin{verbatim}
<|startofname|>Recipe Name<|endofname|>
<|startofingr|>
  Ingredient 1; Ingredient 2; ...;
  Ingredient n
<|endofingr|> 
<|startofinst|>
  Instruction 1. Instruction 2. ...
  Instruction m.
<|endofinst|>
\end{verbatim}
  \caption{Recipe format used to fine-tune \mbox{GPT-2}.  Special tokens define the start and end of each recipe section.}
  \label{fig:recipe-format}
\end{figure}

For evaluation purposes, we split the recipe generation problem into two separate tasks.
Figure~\ref{fig:nameingrtask} shows the \nameingr~task where we prompt the system with the recipe name and ask the system to generate a list of ingredients.
Figure~\ref{fig:nameingrinst} shows the  \nameingrinst~task where we prompt the system with the recipe name and ingredients, and ask the system to generate the recipe instructions.
We have split the task in this manner to simplify the evaluation of the two distinct types of structured text present in recipes.
Note that running these two tasks one after another allows us to generate complete recipes from just the recipe name.

\begin{figure}[t]
\textbf{Prompt:}
\begin{verbatim}
<|startofname|>
Chocalate Chip Cookies
<|endofname|> 
\end{verbatim}
\textbf{Response:}
\begin{verbatim}
<|startofingr|>
1/2 cup butter; 1/2 cup sugar;
1 large egg; 2 cups all-purpose flour;
1 cup semi-sweet chocolate chips;
<|endofingr|>
\end{verbatim}
\caption{In the \nameingr~task the system is given a recipe name and asked to generate a list of ingredients.}
\label{fig:nameingrtask}
\end{figure}

\begin{figure}[t]
\textbf{Prompt:}
\begin{verbatim}
<|startofname|>
Chocalate Chip Cookies
<|endofname|> 
<|startofingr|>
1/2 cup butter; 1/2 cup sugar;
1 large egg; 2 cups all purpose-flour;
1 cup semi-sweet chocolate chips;
<|endofingr|>
\end{verbatim}
\textbf{Response:}
\begin{verbatim}
<|startofinst|>
Preheat oven to 350F.  Combine all 
ingredients in mixing bowl.  Mix in
chocolate chips. Place on baking sheet
and bake for 10 minutes.
<|endofinst|>
\end{verbatim}
\caption{In the \nameingrinst~task the system is given a recipe name and a list of ingredients.  The system is then asked to generate cooking instructions.}
\label{fig:nameingrinst}
\end{figure}
LLMs have a tendency to repeat text, especially the previous sentence at any point.
These repetitions have a self-reinforcing effect --- the probability of repeating a sentence successively increases with each repetition \cite{Xu2022LearningGeneration}.
Penalizing repetitions during inference can help mitigate this problem to some extent.
Two common methods for inference time mitigation include (i) strictly disallowing n-gram repetitions and (ii) exponentially penalizing token repetitions \cite{ShirishKeskar2019CTRL:Generation}.
Both these methods modify the token probabilities of LLMs during inference.
A second major limitation of LLMs is their inability to reliably model long-range dependencies which leads to inconsistent text.
For example, ingredients mentioned in the ingredients list may never be used in the generated recipe instructions (see Figure \ref{fig:InconsitentRecipes}).
Indiscriminately penalizing outputs for repetitions can further aggravate this problem because some repetitions, such as ingredients in recipes, naturally arise owing to their structure.

\subsection{Monte Carlo Tree Search}
MCTS is a search algorithm commonly used in AI agents for playing strategy games such as Chess, Go, and Checkers \cite{Swiechowski2022MonteApplications}.
It balances exploitation and exploration to efficiently search a large search space to find the path that maximizes a user-provided reward function.
In text generation, MCTS can take a long-term view of the text generation process because it evaluates multiple possible paths emanating from the text prompt \cite{Chaffin2022PPL-MCTS:Decoding}.
The algorithm works by maintaining a partial search tree of all possible sentences as shown in Figure~\ref{fig:main}.
The root node $r$ of the tree represents the initial prompt $x_{1:r}$.
The children of each node represent the possible continuations of the current sentence.
\mbox{MCTS} starts with just the root node and then expands one node in each iteration of the algorithm.
The figure shows the state of the algorithm after two iterations have been completed, and the root and ``together'' nodes have been expanded.
The figure also demonstrates how the third iteration proceeds.
In each iteration, \mbox{MCTS} performs the following main steps:

\begin{figure}[t]
  \centering
  \includegraphics[width=0.9\columnwidth]{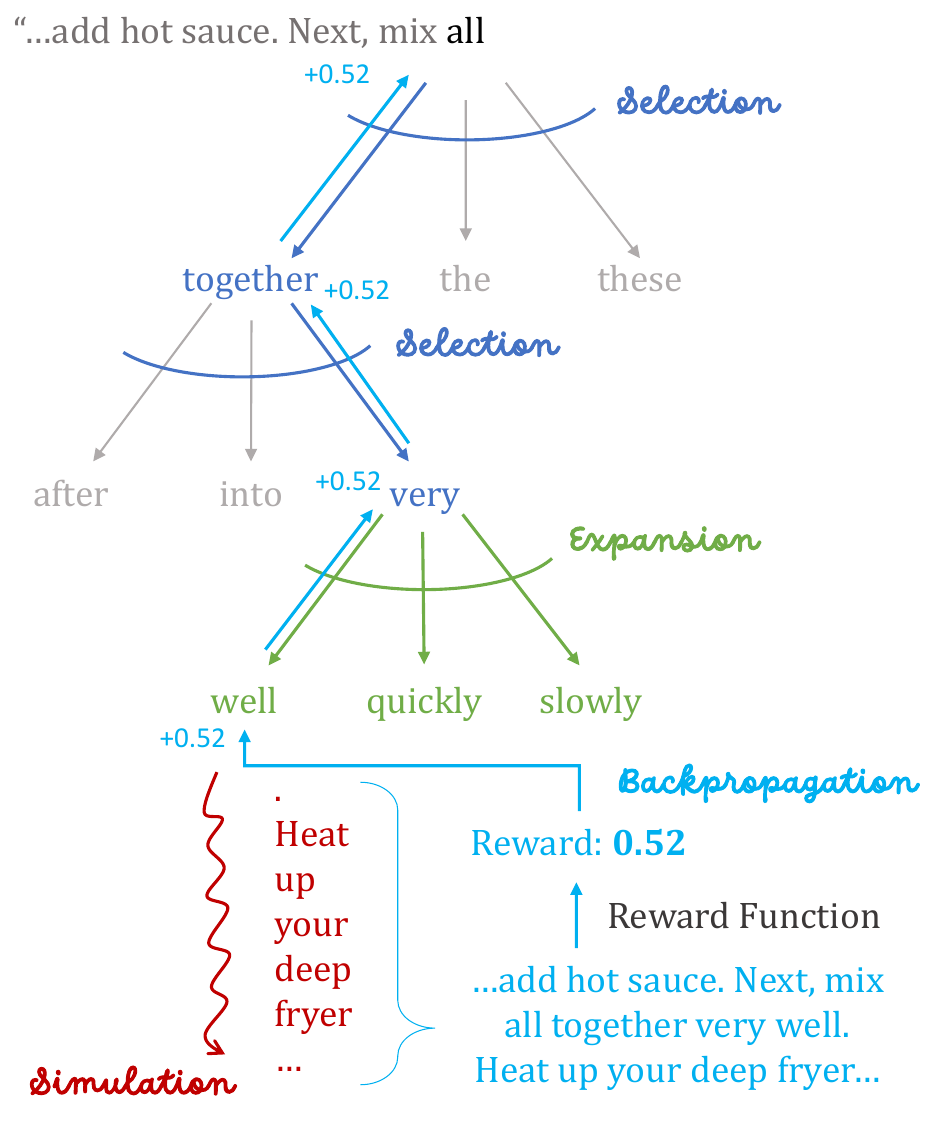}
  \caption{An illustrative example of \mbox{RecipeMC} in action for generating instructions. The initial prompt is ``...add hot sauce. Next, mix all''. The four \mbox{MCTS} steps \textcolor[HTML]{4472C4}{Selection}, \textcolor[HTML]{70AD47}{Expansion}, \textcolor[HTML]{C00000}{Simulation} and \textcolor[HTML]{00B0F0}{Backpropagation} are shown and color coded.}
  \label{fig:main}
\end{figure}

\begin{enumerate}
    \item \textbf{Selection:} Starting from the root node, \mbox{MCTS} iteratively selects a child node to explore until it reaches a leaf node which will be expanded in the next step.
    The node selection is based on maximizing a variant of PUCB (Predictor + Upper Confidence Bound) \cite{Rosin2011Multi-armedContext,Silver2017MasteringKnowledge} over the child nodes:

    $$ \textrm{PUCB}(i) = Q(i) + c \cdot p(x_i|x_{1:i-1}) \frac{\sqrt{N}}{n_i+1}$$

    where the exploitation term $Q(i)$ is the average score from generating token $x_i$ given $x_{1:i-1}$.
    $Q(i)$ is initialized with $0$ whenever a node is created in the expansion step.
    It is updated for all selected nodes during backpropagation.
    The second term, $c$, is a constant that controls the weight of exploration and exploitation; 
    $n_i$ is the count of times that the child node $i$ was visited; and $N=\sum_i n_i$ is the total number of iterations.
    The term $p(x_i|x_{1:i-1})$ is the predictor probability that serves as a prior.  \mbox{RecipeMC} uses the output probabilities from the fine-tuned \mbox{GPT-2} model for $p(\cdot)$.
    Note that PUCB$(i)$ is also well-defined for unexplored child nodes that have $n_i=0$.
    In Figure \ref{fig:main}, the selection algorithm first picks ``together'' because it maximizes PUCB among the three options. It then selects ``very'' because it also maximizes \mbox{PUCB}.  The selection step ends at ``very'' as it is a leaf node.

    \item \textbf{Expansion:} From the selected leaf node $l$, we expand the tree by adding child nodes corresponding to the top-$k$ tokens predicted by the fine-tuned \mbox{GPT-2} model.  We initialize the prior for each child by normalizing the probabilities over the top-$k$ tokens to sum to one.
    In Figure \ref{fig:main}, three nodes corresponding to the words ``well'', ``quickly'' and ``slowly'' are added.
    We used top-$k$ here to limit the tree size as top-$p$ will lead to an indeterministic size.

    \item \textbf{Simulation:} We perform standard top-$p$ sampling from the selected leaf node to generate the next $t$ tokens giving us a text sequence $x_{1:l+t+1}$.
    In Figure \ref{fig:main}, first, the leaf node ``well'' is selected using \mbox{PUCB}.  Then, top-$p$ sampling is used from ``well'' to generate the sequence ``[period] Heat up your deep fryer [...].'' 
    
    \item \textbf{Backpropagation:} We calculate the reward for text sequence $x_{1:l+t+1}$ and update aggregate scores $Q(\cdot)$ and increment $n_{(\cdot)}$ starting from node $l+1$ to the root node $r$.
    In Figure \ref{fig:main}, the scores $Q(\cdot)$ are updated to accumulate a reward of $0.52$, and $n_{(\cdot)}$ is incremented for the words ``well'', ``very'', ``together'', and the root node.
\end{enumerate}

Finally, after repeating the above four steps $Z$ times, one can decide the next token at the root node by choosing the child node with the highest $Q(r+1)$ or the node with the highest $n_{r+1}$.
In our work, we select the node with the highest $Q(r+1)$ as the next token at the root.
We used $Z=20$, $c=1$, $k=50$, $p=0.9$, and $t=30$ without any further fine-tuning.
To generate ingredients or instructions, this process is repeated until the corresponding end tag is reached.

\begin{table*}[t]
  \centering
  \caption{
    Automatic evaluation results for \nameingr~task using different sampling methods. The down arrow ($\downarrow$) indicates that lower is better.
  }
  \vspace{7pt}
  \label{tab:results-ingr}
  \renewcommand{\arraystretch}{1.1}
  \begin{tabular}{lccccccc}
     \hline
    \textbf{Sampling Method} &  \textbf{Coherence} &     \textbf{$F_1$-Score} &  \textbf{Perplexity$\downarrow$} &    \textbf{ROUGE-1} &   \textbf{ROUGE-2} &   \textbf{BLEU} &  \textbf{Repetition$\downarrow$} \\
    \hline
    Ground Truth &              0.451 &         - &         2.934 &         - &         - &         - &     0.667 \\
    \hline
    Top-$p$ &                   0.443 &         0.572 &     4.173 &         0.457 &     0.200 &     0.155 & 1.724 \\
    ~~+ No 4-gram Repetition &  0.444 &         0.562 &     5.150 &         0.456 &     0.198 &     0.144 & 1.641 \\
    ~~+ Repetition Penalty &    0.413 &         0.548 &     6.754 &         0.407 &     0.135 &     0.115 & 0.711\\
    RecipeMC &                  \textbf{0.513}& \textbf{0.597} &     \textbf{3.961}& \textbf{0.505} & \textbf{0.242} & \textbf{0.210} & \textbf{0.192} \\
    \hline
  \end{tabular}
\end{table*}

\subsection{Reward Functions}
While previous \mbox{MCTS} work on \mbox{LLMs} has used discriminator models to guide the MCTS search \cite{Chaffin2022PPL-MCTS:Decoding}, we use hand-designed soft constraints implemented by simple reward functions. 

The reward functions discussed below use a predefined list of common ingredients such as milk, eggs, butter, chicken, etc.\ which we call \emph{constituents}.
To create the constituents list, we applied the NYT Ingredient Phrase tagger\footnote{https://github.com/nytimes/ingredient-phrase-tagger} (NYT-IPT)
over the ingredient phrases in the \mbox{RecipeNLG} dataset.
The \mbox{NYT-IPT} allows tagging of quantities, units, ingredient names, and comments within ingredient phrases.
We extracted the set of ingredient names from all ingredient phrases in the ingredients lists and filtered them to remove any constituents with non-alphabet characters or stop words (``and'', ``or'', etc.).
Further, we filtered out constituents that could be decomposed into other constituents present in the list.
The final list has 2,122 constituents.

Since \nameingr~and \nameingrinst~tasks have very different outputs, they require different reward functions.
The reward functions discussed below were designed to address several structural shortcomings we found in recipes that were generated without \mbox{MCTS}.
For instance, several generated recipes without \mbox{MCTS} failed to include the key ingredient that defines the recipe such as not including chicken as an ingredient in ``Chicken Masala.''
In \mbox{RecipeMC}, we used the following three reward functions for \nameingr~task:

\begin{itemize}
    \item \textbf{Name \& Ingredients Coherence:} This function rewards the model for using ingredients names present in the recipe name.
    For example, for the recipe \textit{Chocolate Apple Pie}, the function rewards outputs with ingredients \textit{Chocolate} and \textit{Apple}.
    We first search for constituents in the recipe name.
    If $z>0$ constituents are found, we search these $z$ constituents in the generated ingredients list and, say, $z_f$ are found.
    The reward value is $z_f/z \in [0,1]$ if $z>0$, and $1$ otherwise.
    
    \item \textbf{Constituents Repetition Penalty:} This function penalizes any repetition of constituents in the ingredient list. Let $p$ be the sum of the number of times a constituent is repeated. Note that we do not count the first occurrences. Similarly, let $q$ be the sum of the number of times the ingredient phrases separated by ``;'' are repeated. Then, the reward is given by $e^{-p-q} \in (0,1]$.
    \item \textbf{Closing Ingredients List:} This function rewards the \texttt{<|endofingr|>} token generation. If the tag is found, the reward is $1$, and $0$ otherwise.
\end{itemize}

The total reward $q$ is defined as the weighted sum of the reward functions ($q = \sum_i w_i r_{i}$) where weight $w_i \in (0,1)$ is assigned such that $\sum_{i} w_i = 1$. This ensures $ q \in [0,1]$ since each function value $r_{i} \in [0,1]$.
We used the weights 0.30, 0.45, and 0.25 for the above three functions respectively without any hyper-parameter fine-tuning.

Similarly, we also use three reward functions for the \nameingrinst~task:

\begin{itemize}
    \item \textbf{Ingredients \& Instructions Coherence:} Similar to \textit{Name \& Ingredients Coherence}, this function rewards the use of constituent names present in the ingredients list when generating recipe instructions. We first search for constituents in the ingredients list. If $z>0$ constituents are found, we search these $z$ constituents in the generated instructions (say $z_f$ are found). The reward value is $z_f/z \in [0,1]$ if $z>0$, and $1$ otherwise.
    \item \textbf{Special Characters Repetition Penalty}: We observed the model tends to repeat some characters like ``!'' and ``-'' because they may be repeated in some training recipe instructions. If the sum of occurrences of these characters is $s$, the reward is given by $e^{-s/S} \in (0,1]$. We used $S=3$ to avoid excessive penalization for using these characters.  
    \item \textbf{Closing Recipe Instructions:} This functions rewards the \texttt{<|endofinst|>} tag. If the tag is found, the reward is $1$, and $0$ otherwise. 
\end{itemize}

For generating instructions, we used the weights 0.50, 0.20 and 0.30 for the above three functions respectively without any hyper-parameter fine-tuning.

\section{Experiments and Results}

We create a LLM for our experiments by fine-tuning \mbox{GPT-2} on our cleaned \mbox{RecipeNLG} data.
We use the same LLM for all our experiments.
We separately evaluate \mbox{RecipeMC} on the \nameingr~task and the \nameingrinst~task.
We compare \mbox{RecipeMC} with following three baseline sampling methods commonly used with LLMs:

\begin{itemize}
    \item \textbf{Top-$p$ Sampling:} Top-$p$ sampling, also known as nucleus sampling \cite{Holtzman2020TheDeGeneration}, uses tokens with the highest probabilities that cumulatively add up to the nucleus size $p$ and zeroes out the probability of other tokens.
    This is an adaptive version of top-$k$ sampling where exactly $k$ tokens with the highest probabilities are considered independent of their cumulative probability. %
    This method has been shown to generate more diverse and interesting text than other greedy approaches. 
    
    \item \textbf{Top-$p$ Sampling with Repetition Penalty:} To prevent the repetition of tokens, the output logit values of repeated tokens are divided by a parameter $\theta>1$ and the distribution is re-normalized\cite{ShirishKeskar2019CTRL:Generation}.
    We use the recommended value $\theta=1.2$ along with top-$p$ sampling as a baseline.
    
    \item \textbf{Top-$p$ Sampling with No $n$-gram Repetitions}:
    This baseline method also  uses top-$p$ sampling but forbids repetition of $n$-grams.
    To ensure that no $n$-gram is repeated, we can search for the last $n$-1 generated tokens in the sequence generated so far and find the list of tokens that follow them.
    These tokens should not be generated to ensure that there are no $n$-gram repetitions.
    This method enforces a strict constraint unlike the repetition penalty but it allows repetitions of $n$-$1$-grams or smaller sequences without any penalization.
    We used $n=4$ in our experiments to allow for some margin in repetition. 
    
\end{itemize}

\begin{table*}[t]
  \centering
  \caption{Automatic evaluation results for \nameingrinst~task using different sampling methods. The down arrow ($\downarrow$) indicates that lower is better.}
  \vspace{7pt}
  \label{tab:results-inst}
  \renewcommand{\arraystretch}{1.1}
  \begin{tabular}{lccccc}
    \hline
    \textbf{Sampling Method} &  \textbf{Coherence} & \textbf{Perplexity$\downarrow$} &    \textbf{ROUGE-1} &   \textbf{ROUGE-2} &   \textbf{BLEU} \\
    \hline
    Ground Truth &              0.486 &     4.115 &         - &         - &         - \\
    \hline
    Top-$p$ &                   0.709 &     7.948 &         0.338 &     0.102 &     0.067 \\
    ~~+ No 4-gram Repetition  & 0.690 &     8.441 &         0.339 &     0.103 &     0.069 \\
    ~~+ Repetition Penalty  &   0.416 &     11.680 &        0.301 &     0.072 &     0.044 \\
    RecipeMC  &         \textbf{0.768} &    \textbf{7.337} & \textbf{0.362} & \textbf{0.115} & \textbf{0.080} \\
    \hline
  \end{tabular}
\end{table*}

\subsection{Automatic Evaluation}

We compare \mbox{RecipeMC} with the three baselines on several standard metrics.
We used 1,000 test recipes with ground-truth ingredient lists and instructions and compare the generated ingredient lists and instructions 
for each method with the ground truth.
For some metrics where ground truth is not required, we also report the values for ground truth as an oracle reference.
For the \nameingr~task, we gave the recipe name from each test recipe as a prompt and sampled the LLM's output ingredients list using all three baseline methods and RecipeMC till the model generated the end tag.
Similarly, for the \nameingrinst~task, we gave the recipe name and ingredients from each test recipe as a prompt and sampled the output instructions.
The results from the automatic evaluation for \nameingr~and \nameingrinst~tasks are summarized in Table \ref{tab:results-ingr} and \ref{tab:results-inst} respectively.

\textbf{Coherence:}
We compare the \emph{coherence} of different methods by comparing constituents present in the generated ingredients list and the recipe name, and between instructions and the ingredients list.  This definition is inspired by the definition of coherence given
by \citeauthor{Lee2020RecipeGPT:System}~\shortcite{Lee2020RecipeGPT:System}.
For \nameingr~task, we define coherence with \textit{Name \& Ingredients Coherence} function defined earlier.
Similarly, for \nameingrinst~task, we define coherence with the \textit{Ingredients and Instructions Coherence} function.
The results in Tables \ref{tab:results-ingr} and \ref{tab:results-inst} show that RecipeMC achieves the highest coherence, surpassing even the ground-truth recipes.
These results confirm that the Coherence reward functions have the desired effect of improving the coherence of the recipes generated with MCTS.

\textbf{$F_1$-Score:} (only for \nameingr~task)
In order to compare the quality of generated ingredients with respect to ground truth, we search for constituents in the ground-truth ingredients list and calculate the average precision and recall for the generated ingredients list.
The $F_1$-Score is the harmonic mean of average precision and recall.
The results in Table \ref{tab:results-ingr} show that RecipeMC has the highest $F_1$-score among all methods confirming that the \textit{Name \& Ingredients Coherence} reward also leads to higher ingredients accuracy with respect to the ground truth.

\textbf{Perplexity:}
The perplexity of a sample, defined with respect to a language model, measures the surprise of the model in seeing the given example.
It is defined as the exponent of cross-entropy over the sequence of tokens in a given text.
Even though LLMs are trained to minimize the cross-entropy loss, the text-generation process or the inference method can influence the perplexity of a sampled text.
For this metric, we only consider the perplexity of the generated part of the output and mask the output probability for the prompt text.
The result in Tables \ref{tab:results-ingr} and \ref{tab:results-inst} show that RecipeMC generates recipes with the lowest perplexity, but still more than that of the test dataset.
This improvement over other baseline methods is because MCTS allows us to look ahead before the next token generation and to avoid tokens that later lead to poor outputs with lower reward values.
It is interesting to note that our method did not explicitly reward lower perplexity, but its low perplexity is a consequence of the soft constraints imposed by the reward functions and the additional search performed by \mbox{MCTS}.

\textbf{ROUGE \& BLEU:}
ROUGE or Recall-Oriented Understudy for Gisting Evaluation \cite{Lin2004ROUGE:Summaries} is a set of metrics to evaluate the quality of text in summarization and machine-translation tasks.
It measures the quality of output text by measuring the overlap, i.e. recall, precision, and accuracy of n-grams between the output text and reference texts.
We use ROUGE-1 and ROUGE-2 $F_1$ values to measure unigram and bigram overlap between the generated recipe texts and the original recipe texts.
BLEU or Bilingual Evaluation Understudy \cite{Papineni2002BLEU:Translation} metric was proposed to measure the quality of machine-translation systems and has been shown to have a high correlation with human judgment.
It combines the precision of $n$-grams where $n=1,2,3,4$, and a brevity penalty for generating output text shorter than reference text.
We measure the BLEU score of output ingredients and instructions by comparing them to the original ones in the test set.
The results in Tables \ref{tab:results-ingr} and \ref{tab:results-inst} show that RecipeMC generates recipes closest to the ground-truth recipes.
Note that the same GPT-2 model was used for each method, but the sampling process used by RecipeMC led to higher-quality ingredient lists and instructions.

\textbf{Repetition:} (only for \nameingr~task)
Repetition is defined as the average number of repetitions of constituents in the ingredients list.
Zero value indicates that no constituents were repeated in the ingredients list.
Table \ref{tab:results-ingr} shows that RecipeMC leads to minimum repetitions, but the repetition value for RecipeMC is even lower than that of the ground-truth recipes.
This may indicate that we are over-penalizing the constituent reuse in the ingredients list.

\textbf{Output Length:}
The average character length of output ingredient lists and instructions are reported in Table \ref{tab:output-lengths}.
We observe that ground-truth recipes have the shortest length.
RecipeMC has the shortest length among all sampling methods since we reward the generation of the end tag.
It is interesting to note that top-$p$ sampling with repetition penalty has the shortest length among the baseline methods for ingredient lists but a much higher length than the other baselines for instructions.
We observed on inspection that the repetition penalty led to a quicker generation of the end tag for ingredients but created a  \textit{blabbering} effect for instructions where the model generates extremely elaborate instructions and occasional unrelated text about alternate or complementary recipes using some ingredients that were not present in the given ingredients list.

\begin{table}[t]
    \caption{
        Average character length of the ingredients list and instructions for different sampling methods.  
    } 
    \vspace{7pt}
    \label{tab:output-lengths}
    \centering
    \renewcommand{\arraystretch}{1.1}
    \begin{tabular}{lcc}
        \hline
        \textbf{Method} & \textbf{Ingredients} & \textbf{Instructions} \\
        \hline
        {Ground Truth} & 167 & 240 \\ 
        \hline
        Top-$p$  & 247 & 485 \\
        ~~+ No 4-gram Repetition  & 248 & 484  \\
        ~~+ Repetition Penalty & 233 & 545 \\
        RecipeMC & {190} & {441} \\
        \hline
    \end{tabular}

\end{table}

Overall, RecipeMC outperformed the baseline methods while balancing several objectives which led to good-quality recipes.
Using top-$p$ sampling without any constraints outperformed other baseline methods except for its highly repetitive text as constituents were repeated 1.7 times on average as compared to 0.7 times for ground truth.
The top-$p$ sampling baselines with No 4-gram Repetition and Repetition Penalty led to lower repetitions as compared to top-$p$ sampling, but it reduced coherence, $F_1$-Score, ROUGE, and BLEU values and increased perplexity.
RecipeMC achieved the best of both worlds with higher $F_1$-score, ROUGE, and BLEU values when compared to ground-truth recipes with the least repetitive text.
The smaller average length of output for RecipeMC (Table \ref{tab:output-lengths}) also confirmed that it was able to succinctly capture more relevant information.
As discussed next, our human evaluations also confirmed that humans prefer recipes generated by RecipeMC.

\subsection{Human Evaluation}

To evaluate each method on \nameingr~task using human evaluation, we created a \textit{Recipe Turing Test} where a human evaluator looks at two possible ingredient lists, the real one and the generated one, for a given recipe name, and their task is to \textbf{choose the generated ingredients list} among these.
For a fair comparison, we uniformly shuffled the real and generated recipes to show them on the left or right side.
Also, evaluators were not aware that four different methods were being evaluated.
We perform a similar test for \nameingrinst~task where human evaluators are asked to \textbf{choose the generated instructions} based on a recipe name and the associated ingredients list.

We randomly sampled 50 recipes for each of the four methods and created a total of 200 binary-choice questions.
The evaluators see 10 randomly-chosen questions (without replacement) for \nameingr~task and 5 questions for \nameingrinst~task on a separate test.
Note that all evaluators did not take both the tests.
The results of the human evaluation for the two tasks are shown in Tables \ref{tab:human-ingr} and \ref{tab:human-inst}.
\textit{Real} and \textit{Gen.} columns count the number of times the real and generated recipes were selected by the evaluator.
$P$(Incorrect) is the probability that humans incorrectly identified the real recipe as the generated one.
This is calculated as follows:

\[P\textrm{(Incorrect)} = \frac{\#\textrm{Real}}{\#\textrm{Real} ~+~ \#\textrm{Generated}}.\]

\begin{table}[t]
    \caption{
        Human evaluation results on the \nameingr\ task.  Data was collected from 147 evaluators who answered 10 questions each.
        Evaluators were asked to identify the generated ingredients (\textbf{Gen.}), hence $P$(\textbf{Incorrect}) = $P$(\textbf{Real}) = \#\textbf{Real} / (\#\textbf{Real} + \#\textbf{Gen.}). 
    } 
    \vspace{7pt}
    \label{tab:human-ingr}
    \centering
    \renewcommand{\arraystretch}{1.1}
    \begin{tabular}{lccc}
        \hline
        \textbf{Method} & \textbf{Real} & \textbf{Gen.} & \textbf{$P$(Incorrect)} \\
        \hline
        Top-$p$ & 175 & 185 & 0.4861 \\
        ~~+ No 4-gram Repetition & 179 & 200 & 0.4723 \\
        ~~+ Repetition Penalty & 183 & 180 & 0.5041 \\
        RecipeMC & 201 & 167 & \textbf{0.5462} \\
        \hline
        \textbf{Overall} & 738 & 732 & 0.5020 \\
        \hline
    \end{tabular}

\end{table}

\begin{table}[t]
    \caption{
        Human results on the \nameingrinst\ task.  Data was collected from 83 evaluators who answered 5 questions each.
        Evaluators were asked to identify the generated ingredients (\textbf{Gen.}), hence $P$(\textbf{Incorrect}) = $P$(\textbf{Real}) = \#\textbf{Real} / (\#\textbf{Real} + \#\textbf{Gen.}).
    }  
    \vspace{7pt}
    \label{tab:human-inst}
    \centering
    \renewcommand{\arraystretch}{1.1}
  
    \begin{tabular}{lccc}
        \hline
        \textbf{Method} & \textbf{Real}& \textbf{Gen.} & \textbf{$P$(Incorrect)} \\
        \hline
        Top-$p$ & 51 & 42 & 0.5484 \\
        ~~+ No 4-gram Repetition & 67 & 62 & 0.5194 \\
        ~~+ Repetition Penalty & 36 & 65 & 0.3564 \\
        RecipeMC & 57 & 35 & \textbf{0.6196} \\
        \hline
        Overall & 211 & 204 & 0.5084 \\
        \hline
    \end{tabular}
\end{table}

The results in Tables \ref{tab:human-ingr} and \ref{tab:human-inst} show that human evaluators are most likely to believe that \mbox{RecipeMC} ingredient lists and instructions are human-generated when compared to those produced by the other baselines.
Human evaluators believed recipes generated by \mbox{RecipeMC} to be human-generated more often than the next best method, top-$p$ sampling with Repetition Penalty, for \nameingr~task with a $p$-value of $0.128$.
Similarly, \mbox{RecipeMC} outperforms top-$p$ sampling on the \nameingrinst~task with $p=0.164$.
While we did not observe statistically significant ($p < 0.05$) improvements with \mbox{RecipeMC} results, we observe that results from human evaluation correlate well with results from automatic evaluation presented in Tables \ref{tab:results-ingr} and \ref{tab:results-inst}.

We also observed that top-$p$ sampling with repetition penalty performed at par with other baselines for \nameingr~task but performed worse than other baselines on the \nameingrinst~task with $p=0.007$.
This can be attributed to its unusually long instructions as discussed earlier and shown in Table \ref{tab:output-lengths}.

\remove{We also note that human evaluators prefer RecipeMC more often than the original recipes in both experiments.}
\new{
Juxtaposing these results with average lengths shown in Table \ref{tab:output-lengths}, we observe that human evaluators prefer recipes with shorter length among the generative methods, but do not prefer the shortest recipes, i.e the real recipes, over RecipeMC for both ingredients and instructions.
This confirms that humans did not heavily rely on text length.  
}
Assuming that human evaluators select outputs at random when recipes are equally good, it is surprising to see that the ingredient lists generated by \mbox{RecipeMC} are perceived to be more human-like than the original (human-written) recipes with $p=0.042$ and instructions are perceived to be more human-like than original recipes with $p=0.014$.
We believe that this is because (i) \mbox{RecipeMC} repeats ingredients far less than the original recipes, as shown in Table \ref{tab:results-ingr}, and (ii) it uses complete ingredient names consistently, as shown in Figure \ref{fig:InconsitentRecipes}, and indicated by coherence values higher than ground truth shown in Tables \ref{tab:results-ingr} and \ref{tab:results-inst}.

\section{Discussion and Future Work}
\mbox{MCTS} and manually defined reward functions provide an effective way to control the ingredients and instruction generated by \mbox{LLMs} without requiring additional training.
This approach offers flexibility and can enable users to generate recipes that adhere to specific constraints.
It can also be valuable for interactive recipe editing.
Users can present partial recipes to the system and ask the system to fill in the blanks{.
Personal constraints such as being sugar-free, low sodium, or vegetarian can be added to the reward function.
Users can} iteratively prompt the system with different combinations of ingredients and collaboratively create new recipes.
We plan to explore interactive recipe generation in future work \new{to measure novelty and creativity through user studies with amateur and professional chefs}.
\remove{This approach can also be applied to a variety of other structured text problems beyond recipes such as generating test questions, checklists, calendar entries, etc.}

The ability of MCTS to look ahead during the generation of each token coupled with heuristic-based reward functions interestingly led to lower perplexity and higher similarity to ground truth recipes without directly optimizing for these properties.
This finding is interesting from the perspective of text generation and suggests that using MCTS with LLMs may be applicable to a wider set of applications beyond structured text generation.

\section{Conclusions}
We presented a new sampling method, \mbox{RecipeMC}, that combines \mbox{LLMs}, \mbox{MCTS}, and custom reward functions to generate recipes that are often indistinguishable from human recipes for the same dish.
We have shown that our method outperforms common text-generation approaches for \mbox{LLMs} for this task on a variety of automatic generation metrics.
We conducted a \emph{Recipe Turing Test} and found that users preferred \mbox{RecipeMC} ingredients about 55\% of the time and \mbox{RecipeMC} generated instructions 62\% of the time as compared to human-generated ingredients and instructions.
These evaluations show that \mbox{MCTS} combined with manually-defined reward functions can be an effective tool for recipe generation with LLMs such as GPT-2.

\bibliographystyle{iccc}
\bibliography{references}

\end{document}